\documentclass{article}

\PassOptionsToPackage{numbers, compress}{natbib}

\usepackage{microtype}
\usepackage{subcaption}
\usepackage{booktabs} 
\usepackage{graphicx} 
\usepackage{makecell} 
\usepackage{pifont}
\usepackage{xcolor}
\usepackage{bbding}
\usepackage{siunitx} 
\usepackage{arydshln}
\usepackage{algorithm}
\usepackage{wrapfig}

\usepackage{multirow}

\usepackage[hidelinks]{hyperref}

\usepackage{algpseudocode}

\usepackage[final,main]{neurips_2026}

\usepackage{amsmath}
\usepackage{amssymb}
\usepackage{mathtools}
\usepackage{amsthm}
\newtheorem{prop}{Proposition}

\usepackage[capitalize,noabbrev]{cleveref}

\theoremstyle{plain}

\theoremstyle{definition}

\theoremstyle{remark}

\usepackage[textsize=tiny]{todonotes}

\newcommand{\name}{\textsc{Flash PD-SSM}}
\newcommand{\hybridname}{\textsc{Delta-PD}}

\newcommand{\update}[1]{{\color{black} #1}}

\title{\name: Memory-Optimized \\ Structured Sparse State-Space Models}
\author{
Aleksandar Terzić$^{1,2}$\thanks{These authors contributed equally.}\\
{\tt\small aleksandar.terzic1@ibm.com}\\
\And
Francesco Carzaniga$^{1}$\footnotemark[1]\\
{\tt\small frc@zurich.ibm.com}\\
\AND
Nicolas Menet$^{1,2}$\\
{\tt\small nicolas.menet@ibm.com}\\
\And
Yannick Biehl$^{2}$\thanks{Research conducted at IBM Research -- Zurich.}\footnotemark[1]\\
{\tt\small ybiehl@student.ethz.ch}\\
\And
Michael Hersche$^{1}$\\
{\tt\small michael.hersche@ibm.com}\\
\AND
Thomas Hofmann$^{2}$\\
{\tt\small thomas.hofmann@inf.ethz.ch}\\
\And
Abbas Rahimi$^{1}$\\
{\tt\small abr@zurich.ibm.com}
\AND
{
\normalfont $^{1}$IBM Research -- Zurich, $^{2}$Department of Computer Science, ETH Zürich}
}

\begin{document}

\maketitle

\begin{abstract}

State-space models (SSMs) face a fundamental trade-off between efficiency and expressivity that is mainly dictated by the structure of the model's transition matrix. 
Unstructured transition matrices enable maximal expressivity, as measured by their ability to model finite-state automaton (FSA) transitions, but come at a prohibitively high compute and memory cost.
In contrast, most structured transition matrix forms are highly efficient both in runtime and memory consumption, but suffer from limited expressivity.
%
Building on recent work on structured sparse SSMs, we propose \name, a novel SSM that achieves comparable throughput to widely-used structured SSMs with significantly better expressivity guarantees.
\name\ maintains a trainable set of structured sparse matrices, a single one of which is discretely selected at each time-step, enabling FSA expressiveness at the level of unstructured matrices while maintaining the efficiency required for training models at scale.
First, we validate \name\ against a suite of alternative models on synthetic mechanistic and state-tracking tasks, finding that its theoretical expressivity is achieved in practice. 
Second, on multivariate time-series tasks involving sequences of length over 17\,000, we find that \name\ defines a new state-of-the-art (SoTA) accuracy among competing SSM methods.
Finally, we demonstrate that \name\ is an effective drop-in replacement for hybrid LLMs, yielding improvements both in natural language state-tracking and in common language modeling scenarios.
Up to 3B parameters, its training speed matches that of widely used SSMs, with less memory than DeltaNet and Gated DeltaNet.
Code is available at \url{https://github.com/IBM/flash-pdssm}.
\end{abstract}

\section{Introduction}
\label{sec:intro}

State-space models (SSMs)~\citep{gu_efficiently_2022, gupta_diagonal_2022, fu_hungry_2023, smith_simplified_2023, orvieto_resurrecting_2023, gu_mamba_2023, daotransformers, yang2024parallelizing, siems_2025_deltaproduct} offer a scalable alternative to Transformers~\cite{vaswani_attention_2017}, providing distinct advantages such as linear compute scaling relative to sequence length and constant memory consumption during inference. 
Modern LLMs increasingly employ hybrid architectures that combine the strengths of Transformers and SSMs, often utilizing a higher proportion of SSM layers in order to effectively reap their computational benefits~\citep{yang2024parallelizing, ren_2024_samba, de_griffin_2024, waleffe_2024_empirical, lenz_2025_jamba, wu2025transxssm}.


\begin{figure}[t]
    \centering
    \begin{minipage}[c]{0.58\linewidth}
    \includegraphics[width=\linewidth]{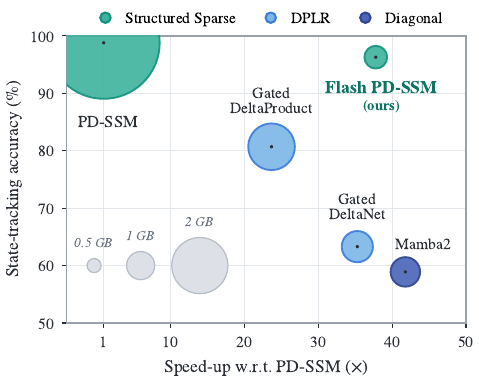}
    \end{minipage}\hfill
    \begin{minipage}[c]{0.39\linewidth}
\caption{\textbf{\name\ is expressive, fast, and memory-efficient.} Synthetic state-tracking accuracy on a collection of FSA emulation tasks~\cite{deletang_neural_2023, walker_2025_structured}. The single-layer runtime (measured relative to PD-SSM~\cite{terzic_2025_pd}) and memory consumption of \name\ and other popular SSMs are also reported. The maximum sequence length is 2048 and all models have hidden dimension 1024. The circle's size indicates peak memory consumption during training. \name\ is the only SSM in the upper right quadrant, i.e., both highly expressive and fast.}
    \label{fig:highlight}
    \end{minipage}
\end{figure}

Initially, the computational costs associated with training large models have required the use of diagonal transition matrices, which impose significant limitations on model expressivity~\citep{gu_mamba_2023, merrill_parallelism_2023, cirone_theoretical_2024, sarrof2024expressivecapacitystatespace}.
Consequently, the design of SSM transition matrices that better balance the expressivity-efficiency trade-off has become an active area of inquiry, spurring a range of works investigating alternative parametrizations~\cite{schlag2021linear, yang2024parallelizing, grazzi2024unlocking, siems_2025_deltaproduct, peng_2025_rwkv, walker_2025_structured, anonymous2025mamba3}. 
Here, we build upon a recent development that utilizes \emph{structured sparse} transition matrices, namely the PD-SSM model~\cite{terzic_2025_pd}. The PD-SSM matrices provide optimal state-tracking expressivity guarantees, at the level of unstructured SSMs, while enabling efficient parallel computation through linear-complexity matrix-matrix multiplication of the highly sparse structured transition matrices.

However, while the structured sparse matrices utilized by PD-SSM offer linear asymptotic complexity in the parallel evaluation of the sequence of hidden states, the generation of the input-dependent matrices incurs a significant overhead. If $L$ is the sequence length and $N$ the SSM hidden state size, generating the sparse transition matrices in PD-SSM incurs an $\mathcal O(LN^2)$ memory cost. In practice, a parameter-matched Mamba2~\cite{daotransformers} is more than $40\times$ faster (see Figure~\ref{fig:highlight}), making PD-SSM unsuitable for large-scale deployment. 
%

In this work, we address this bottleneck by introducing \name, a highly efficient and effective structured sparse SSM. \name\ provides state-of-the-art (SoTA) results in various state-tracking tasks and improves hybrid language models, while maintaining a training throughput comparable with related SSMs widely used in frontier language models, with less memory than DeltaNet~\cite{schlag2021learning, yang2024parallelizing} and Gated DeltaNet~\cite{yanggated}. Our main contributions are as follows.

\textbf{Novel Selection Mechanism and Hardware-Aware Design.} 
First, we design a novel generation mechanism for structured sparse transition matrices that maintains PD-SSM's expressivity guarantees, yet reduces the memory complexity from $\mathcal O(LN^2)$ to $\mathcal O(LN)$.
We combine this algorithmic improvement with a custom hardware-aware Triton/CUDA implementation of the structured linear recurrence of PD-SSM. In parameter-matched language models, \name\ trains up to $1.3\times$ faster than DeltaNet and stays within $6\%$ of it up to 3B parameters (Figure~\ref{fig:lm_efficiency}). 
It uses $4$--$12\%$ less memory than DeltaNet and Gated DeltaNet. Section~\ref{sec:res_runtime} and Appendix~\ref{apx:efficiency} report the full measurements.

\textbf{Synthetic State-Tracking, Mechanistic Capabilities, and Time-Series Modeling.} 
Second, we show that \name\ can achieve the theoretical guarantees of structured sparse SSMs in practice, by exhibiting excellent length generalization accuracy ($96.3\%$ avg.) on synthetic state-tracking tasks~\cite{deletang_neural_2023, walker_2025_structured}, substantially outperforming both diagonal (Mamba2~\cite{daotransformers} with $58.9\%$ avg. and Mamba3~\cite{anonymous2025mamba3} with $73.0\%$ avg.) and DPLR SSMs (Gated DeltaProduct~\cite{siems_2025_deltaproduct} with $80.7\%$ avg.). 
These results confirm the superior state-tracking capabilities of~\name, placing it at the frontier of expressiveness, throughput, and memory usage.

Probing a different type of fundamental model capabilities, on mechanistic capability tests (MAD~\cite{poli2024mechanistic}), \name\ achieves scores comparable to Mamba~\cite{gu_mamba_2023} ($70.4\%$ vs. $69.3\%$). 
%
Since high performance on this task has been linked to strong language modeling capabilities, we use these results to inform our choice of hybridization for the language modeling tasks that follow.

%
%
%
Furthermore, \name\ achieves SoTA results on time-series tasks such as the UEA multivariate time-series classification task~\cite{uea_time_series}, containing time series data of length over 17\,000, and the PPG-DaLiA time-series regression~\cite{ppg_dalia} task.

\textbf{Natural Language State-Tracking and Language Modeling.} 
%
Finally, we show that \name\ is effective in hybrid language models through two classes of tasks: natural language state-tracking and general language modeling. We extend the Box~\citep{kim-schuster-2023-entity} state-tracking dataset to more than twice its original length (up to 26 state transitions) and use it to benchmark frontier hybrid models, Granite4~\cite{granite2025} and NemotronH~\cite{blakeman2025nemotron}, combining self-attention and Mamba2. Here, we demonstrate that \name\ surpasses the out-of-distribution (length generalization) accuracy of Mamba2 by up to $10\%$ in an equal setting while maintaining comparable throughput (Section~\ref{sec:res_runtime}). 

We then consider the pure language modeling capabilities of \name\ to evaluate the feasibility of its inclusion in frontier models. Informed by the results on MAD, we train \textit{\hybridname}, a hybrid language model consisting of \name\ and DeltaNet layers. We pretrain on 20B tokens of FineWeb-Edu~\cite{penedo2024fineweb} and evaluate on common downstream benchmarks~\cite{eval-harness}. Hybrids of \name\ with DeltaNet or attention match or outperform the corresponding pure models in training loss and accuracy (Table~\ref{tab:lm_results}).

\section{Background}
\label{sec:background}


\subsection{State-Space Models}

We use the term \textbf{state-space model} (SSM) to refer to neural networks which implement variants of the following recurrence relations:
\begin{equation} \label{eq_slssm}
x_{t} = A(u_t) x_{t-1} + B (u_t) u_t,\;
y_{t} = C(u_t)x_{t} + Du_t,\;
o_{t} = \psi(y_t)
\end{equation}
where $u_t \in \mathbb{R}^D$ denotes the system input, $x_t \in \mathbb{C}^N$ represents the latent state, and $y_t \in \mathbb{C}^D$ and $o_t \in \mathbb{R}^D$ are the complex- and real-valued outputs, respectively.
The final readout layer $\psi: \mathbb{C}^D \to \mathbb{R}^D$ extracts a real-valued embedding from the complex output, typically implemented as $\psi(y_t) = \text{Re}\{y_t\}$ \citep{gu_parameterization_2022, gupta_diagonal_2022, orvieto_resurrecting_2023}.
The system is parameterized by the state transition matrix $A(u_t) \in \mathbb{C}^{N \times N}$, the input and output projection matrices $B(u_t), C(u_t) \in \mathbb{C}^{D \times N}$, and a weighted skip connection $D \in \mathbb{C}^{D \times D}$. 

\subsection{The Expressivity - Efficiency Trade-Off in SSMs}\label{sec:background_expressiveness}

The most efficient transition matrix structure in SSMs is the scaled identity~\cite{daotransformers} followed by the diagonal matrix~\cite{gu_mamba_2023}. In the linear time-invariant (LTI) case ($A(u_t)=A, B(u_t)=B, C(u_t)=C$), the dynamics can almost always be equivalently represented with a complex diagonal transition matrix, as diagonalizable matrices are dense in the set of square matrices~\cite{orvieto_resurrecting_2023, axler_linear_2024}. 
%

%
However, modern SSMs employed on complex tasks typically utilize time-varying dynamics~\cite{gu_mamba_2023}. In such systems, diagonal transition matrices pose significant limitations on the expressivity~\cite{merrill_illusion_2024, cirone_theoretical_2024, walker_2025_structured}.
Concretely, \citet{merrill_illusion_2024} demonstrate that diagonal time-varying SSMs, both real and complex, are theoretically incapable of emulating FSAs with non-solvable transformation semigroups~\cite{carter_visual_2009, terzic_2025_pd}.
This poses a fundamental limitation on the state-tracking capability of diagonal SSM systems.
%
Higher state-tracking capability in SSMs has been linked to improved perplexity in code and mathematical text generation~\cite{grazzi2024unlocking, siems_2025_deltaproduct}, providing concrete practical motivation for improving such capabilities.
Relaxing the diagonal constraint to allow for arbitrary transition matrices enables fully expressive state-tracking with SSMs, but comes at a prohibitive compute and memory cost~\cite{terzic2025sdssm, merrill_illusion_2024}. 
An alternative approach utilizes (products of) Householder matrices, exemplified by the DeltaNet and DeltaProduct family of models~\cite{schlag2021linear, yang2024parallelizing, siems_2025_deltaproduct}. 
Such matrices increase the FSA expressivity of the resulting SSM, but only provide the theoretical guarantees of unstructured SSMs if given prohibitively wide and deep architectures~\cite{siems_2025_deltaproduct, terzic_2025_pd}.

Another alternative comes in the form of SSMs with structured sparse transition matrices, such as PD-SSM~\cite{terzic_2025_pd}. Within the framework of FSA emulation, such models provide the same theoretical guarantees as unstructured SSMs; namely, they can emulate any FSA with $N$ states using a single layer with hidden size $N$ and a linear readout of size $N\times N$~\cite{terzic2025sdssm, terzic_2025_pd}. 

\subsection{PD-SSM -- SSMs with Structured Sparse Transition Matrices}\label{sec:pdssm}

%

As mentioned previously, PD-SSM circumvents the expressivity limitations of other SSMs by utilizing sparse transition matrices whose structure is preserved under matrix multiplication, a crucial requirement for enabling efficient implementations based on parallel prefix scans~\cite{blelloch_prex_1990}.
Concretely, the PD-SSM matrix generator, denoted by $A(u_t):\mathbb{R}^D \rightarrow \mathbb{C}^{N\times N}$, consists of two main components: a \emph{structured sparse matrix generator} $P(u_t)$ and a diagonal matrix generator $D(u_t)$.
%

\begin{figure}[t]
    \centering
    \begin{subfigure}{\textwidth}
        \centering
        \includegraphics[width=0.84\linewidth]{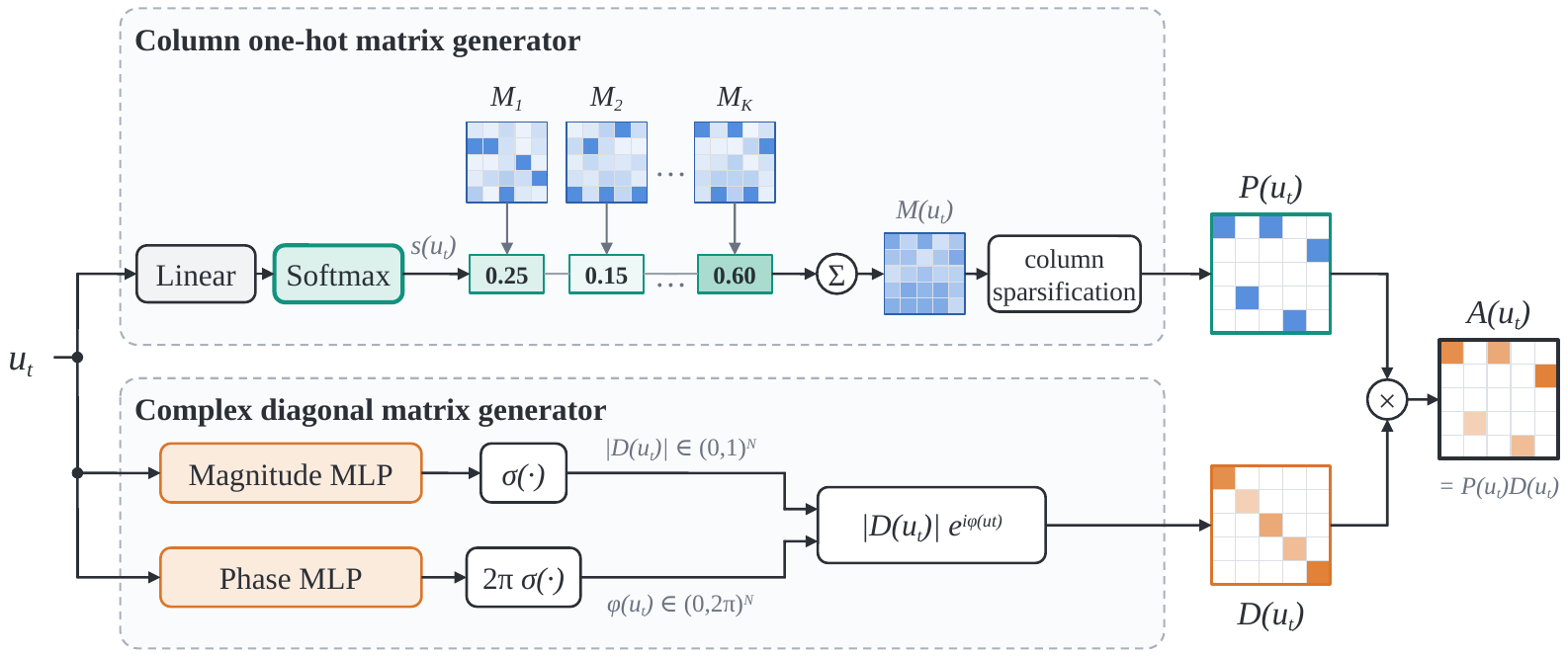}
        \caption{\textbf{PD-SSM}~\cite{terzic_2025_pd} sparsifies a convex combination of dense dictionary matrices. The column one-hot matrix generation process incurs significant computational and memory overheads, as an $N\times N$ matrix $M(u_t)$ is generated at each step.}
        \label{fig:pdssm_sketch}
    \end{subfigure}

    \vspace{0.8em}
    \begin{subfigure}{\textwidth}
        \centering
        \includegraphics[width=0.84\linewidth]{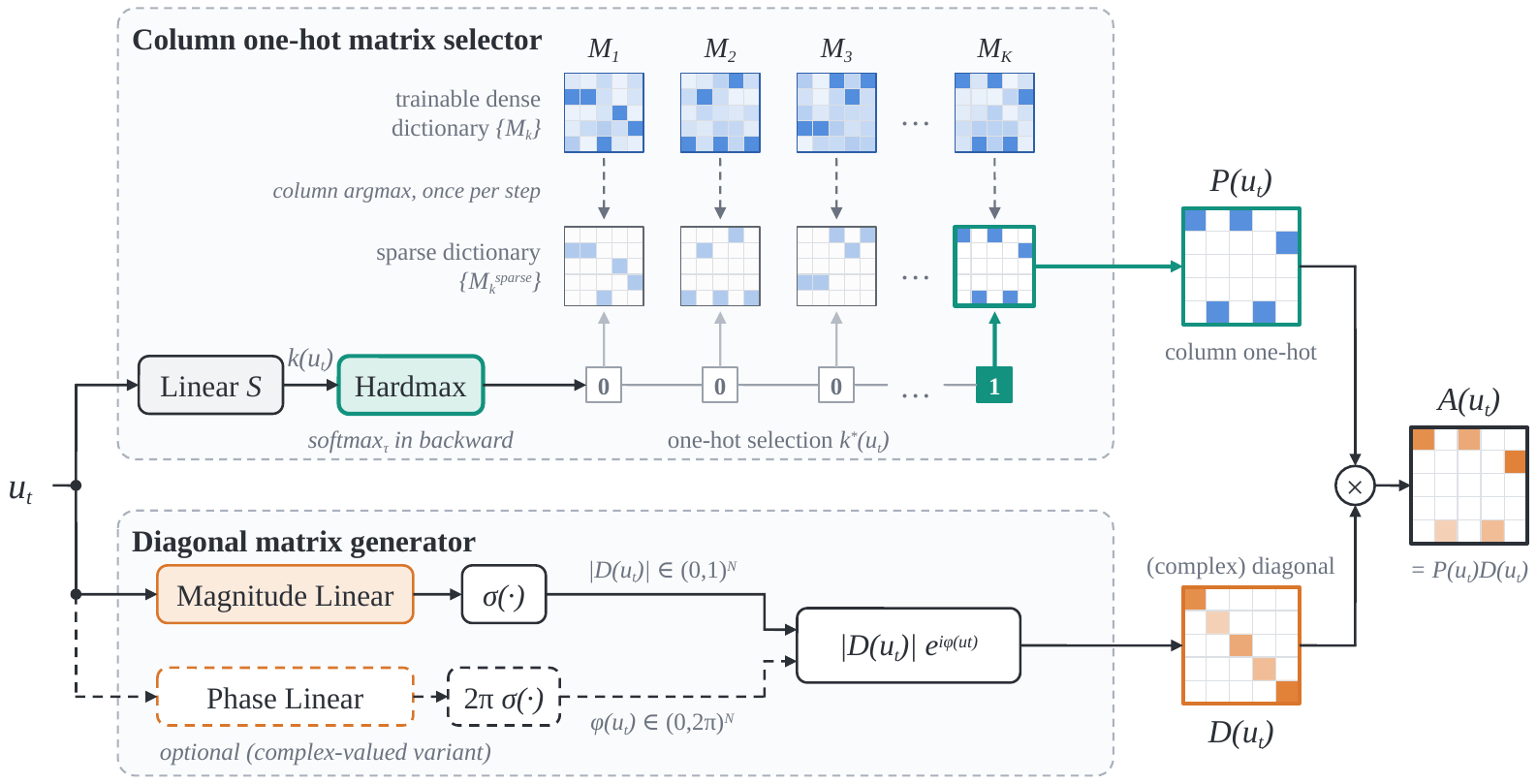}
        \caption{\textbf{\name} simplifies the column one-hot generation process by directly selecting a single element from a dictionary of trainable structured sparse matrices. This preserves the theoretical guarantees and allows for significant runtime improvements.}
        \label{fig:main_model_sketch}
    \end{subfigure}
    \caption{\textbf{Generation of the transition matrices} in PD-SSM (a) and \name\ (b).}
\end{figure}

The crucial component enabling increased expressivity is the structured sparse matrix generator $P(u_t)$. The input $u_t$ generates the weights that are used to soft-select among a set of trainable transition matrices, i.e.,  the dictionary $\{M_i \in \mathbb{R}^{N\times N}\}_{i\in[K]}$ with $K$ being a hyperparameter. 
Sparsity of $P$ is achieved by applying a column-wise hardmax on the resulting matrix, producing a one-hot column matrix. 
This procedure is depicted in Figure~\ref{fig:pdssm_sketch}, with the full equations given by:
\setlength{\jot}{0pt} 
\begin{align*}
    s(u_t) &= \text{softmax}(S u_t) \in \Delta^{K-1} \\
    M(u_t) &= \sum_{k=1}^K s_k(u_t) M_k \in \mathbb{R}^{N \times N} \\
    P_{:,j}(u_t) &= \text{column\_hardmax}(M_{:,j}(u_t)) \in \{0, 1\}^N 
\end{align*}
\setlength{\jot}{3pt} 
%
%

While the resulting sparse $P(u_t)$ matrices are highly efficient for chaining via the parallel scan~\cite{blelloch_prex_1990}, incurring an $\mathcal O(LN)$ computational cost, generating the matrices themselves poses a significant computational and memory bottleneck.
%
%
Materializing the $M(u_t)$ matrices in parallel for sequences of length $L$ incurs an $\mathcal O(L N^2)$ memory footprint, which is prohibitive for large-scale training where both $L$ and $N$ are typically of the order of at least $10^3$.
%

\section{Memory-Optimized \name}
\label{sec:method}


%



\subsection{Novel Sparse Selection Mechanism}\label{sec:sparse selection}

%
%
%
%
In our architectural contribution described in this section, we modify the design of PD-SSM in a manner which reduces the memory footprint by a factor of $N$.
We do this by redesigning the soft-selection mechanism inherent in the computation of $P(u_t)$ to \emph{hard selection}.
Concretely,  \name's new matrix generator $P(u_t)$ is defined as follows:
\begingroup
\setlength{\jot}{0pt} 
\begin{align} 
        M_k^{\text{sparse}}[j] &=\text{argmax}(M_k[:,j]) \in \{0,\dots, N-1\}\label{eq:first_argmax} \\
        k(u_t)&=Su_t \in \mathbb{R}^K \\
        k^*(u_t) &=\text{argmax}(k(u_t)) \in \{1,\dots,K\} \label{eq:second_argmax}  \\
        P(u_t)&=M_{k^*(u_t)}^{\text{sparse}}\in \{0,\dots,N-1\}^{N} 
\end{align}
\endgroup
\vspace{-12pt}
\label{eq:discrete_pdssm}

The sparse matrices $M_k^{\text{sparse}}$ are generated from a dictionary of $K$ dense matrices $M_k, k\in[K]$ by selecting the largest element in each column. This is only done once, e.g. at the beginning of the optimization step, and the resulting sparse matrices are subsequently used in both the forward and backward computation.
At each time-step, the input $u_t$ generates $K$-many values via the linear transformation $S:\mathbb{R}^D\rightarrow \mathbb{R}^K$. The largest of these determines the index of the transition matrix to be selected, denoted by $k^*(u_t) \in[K] := \{1,\dots,K\}$.  The selected matrix is denoted as $M_{k^*(u_t)}^{\text{sparse}}\in[N]^N$. 
%
As each sparse dictionary matrix $M^{\text{sparse}}$ effectively reduces to an array of $N$ integers, and since we now only have to select a single vector of $N$ integers per time step, the memory cost is decreased from $\mathcal O(L N^2)$ to $\mathcal O(LN)$ and the compute cost from $\mathcal O(LN^2 K)$ to $\mathcal O(LNK)$.
%
%
The full \name{} matrix generator is shown in Figure~\ref{fig:main_model_sketch}.

Given any FSA with state size $N$ and input alphabet size $K$, PD-SSM can emulate it by selecting, based on the input, one out of $K$-many $N\times N$ matrices with a single active element per column \cite{terzic_2025_pd}. 
Therefore, instead of sparsifying an input-dependent convex combination of dictionary matrices, we can directly select a sparse transition matrix.
This is explicitly characterized in Proposition~\ref{prop:expressivity}, with proof in Appendix~\ref{sec:apx_proof_expressivity}.

\begin{prop}[Expressivity of Discrete PD Parametrization]\label{prop:expressivity}
Any deterministic FSA with $N$ states can be exactly represented by a single-layer \name{} with a state size $N$ and linear readout of size $N \times N$.
\end{prop}

The model we propose is highly discrete, requiring the use of gradient surrogate techniques.
To propagate gradients, we leverage \emph{slope annealed straight-through estimation}~\cite{bengio_stochastic_2013}.
Concretely, each occurrence of the hardmax operation in the forward pass is approximated via a softmax in the backward pass.
%
%
The straight-through estimator is used twice in the architecture corresponding to the two argmaxes in \cref{eq:first_argmax} and \cref{eq:second_argmax}. For completeness, we state the full gradient expressions in Proposition~\ref{prop:gradients} with the derivation provided in Appendix~\ref{apx:proof}.
%
%



\begin{prop}[Surrogate Gradients]\label{prop:gradients}
    Denote $P_t:= P(u_t)$ and analogously for $D_t$. Let $\partial \ell/\partial\theta$ be the derivative of the loss w.r.t. the parameter $\theta$ and $\text{softmax}_\tau$ be the tempered softmax function. Given $x_t$, $D_t$ and $\partial \ell/\partial x_t$ for each time step, the gradients of the dictionary matrices and the column selector are approximated by the straight-through estimator as
    {\small
    \[
        \frac{\partial \ell}{\partial M_{k}} \approx \sum_{t:k^*(u_t)=k}\Big(\frac{\partial \ell}{\partial x_t}\;(D_t x_{t-1})^T \Big) \frac{\partial \text{softmax}_\tau(M_k)}{\partial (M_k)}
    \]
    \[
        \frac{\partial \ell}{\partial k(u_t)} \approx \big(\frac{\partial \ell}{\partial x_t} (P_tD_tx_{t-1})^T\big)\frac{\partial \text{softmax}_\tau(k(u_t))}{\partial k(u_t)}
    \]}
    The expressions above become exact in the limit of $\tau \rightarrow 0$.
\end{prop}

The discretely selected $P_t$ matrices are reused in the backward pass, i.e., the trainable dense structure of the $M_k$ dictionary matrices does not incur any prohibitive additional overhead in the computation.
The expressions above can be computed in $\mathcal O(LN)$ time and space. We provide custom memory-efficient Triton~\cite{triton} kernels parallelizing the computations. The kernels occupy an additional amount of memory equal to the amount required for the storage of the trainable dense dictionary matrices, this being $O(KN^2)$.
%
%

%
%
%

\begin{figure}[t]
    \centering
    \includegraphics[width=0.9\textwidth]{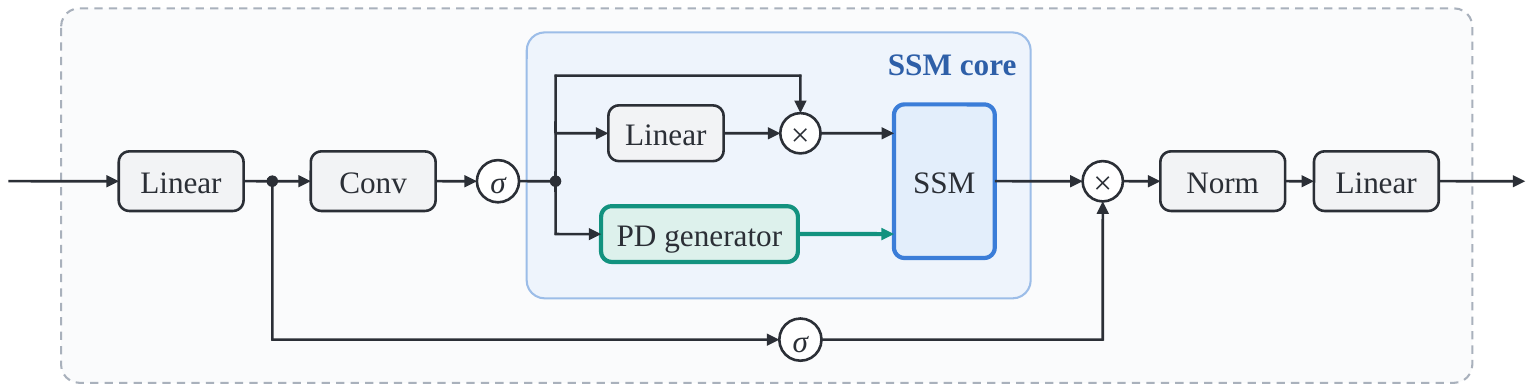}
    \caption{One \name\ block integrates a range of components in a design following the pattern from~\cite{daotransformers}.}
    \label{fig:block}
\end{figure}

\subsection{The \name\ Layer}

We embed \name\ into an interconnected block by following standard design patterns outlined by Mamba~\cite{gu_mamba_2023}, further embedding the block into a standard pre-norm architecture~\cite{pre_post_norm}. The full architecture is shown in Figure~\ref{fig:block}.
%
%
%
%
We adopt a multi-head SSM architecture, which conceptually allows the model to keep track of several independent FSAs simultaneously.
%
The multi-head design allows for an amortization of the parameter cost of the dictionary matrices, which are dense during training.
Consider a model with $D$-dimensional embeddings, $H$ heads of size $D/H$ and $K$ dictionary matrices. If we set the number of heads to $H=\sqrt{D}$, which roughly corresponds to 48 heads when $D=2048$, the total parameter cost of the dictionary matrices becomes $K  D^{3/2}$. If we then set $K=\sqrt{D}$, the parameter cost of the sparse matrix selector becomes equal to that of a single $D\times D$ linear layer without bias.
Therefore, we can scale the number of dictionary matrices as $K=\mathcal O(\sqrt{D})$ without blowing up the parameter cost of the model.

\subsection{Chunk-Wise Implementation}

We design a three-phase CUDA kernel with additional Triton wrappers that efficiently exploit the parallel hardware of GPUs to compute the \name\ recurrence. 
%
%
%
%
The input is partitioned into chunks, and in each chunk, we compute an independent sequential recurrence in parallel, followed by inter-chunk aggregation and another independent sequential per-chunk recurrence that accounts for the additive effects of the inter-chunk aggregation. 
An overview of our method for computing the sequence of states $x_t$ can be found in Appendix~\ref{sec:apx_implementation}.

\section{Experimental Results}\label{sec:results}

\subsection{Runtime Measurements}\label{sec:res_runtime}

\begin{figure}[!htb]
\centering
\includegraphics[width=\textwidth]{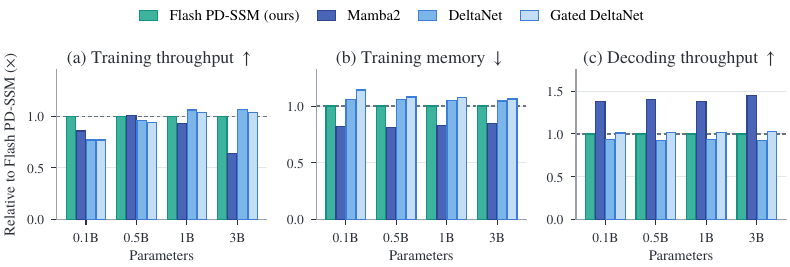}
\caption{\textbf{Language-model efficiency}, relative to \name, at training length 2048: training throughput (a), peak training memory (b) and decoding throughput at batch size 16 (c).}
\label{fig:lm_efficiency}
\end{figure}

We measure parameter-matched language models built from a single mixer type, from 0.1B to 3B parameters (setup and numbers in Table~\ref{tab:lm_efficiency}). \name\ trains fastest at 0.1B and stays within $6\%$ of DeltaNet up to 3B, using $4$--$12\%$ less memory than DeltaNet and Gated DeltaNet. It decodes $7$--$8\%$ faster than DeltaNet (Figure~\ref{fig:lm_efficiency}).

\subsection{Synthetic Expressivity Evaluation of \name}\label{sec:exp_eval}

We first evaluate the algorithmic capabilities of our method using synthetic state-tracking and memory-based tasks to verify that the  expressivity guarantees of \name\ are achieved in practice.

\paragraph{Synthetic State-Tracking.}\label{sec:sst}
We use the state-tracking tasks originally introduced by~\citet{deletang_neural_2023}, comprising of 4 finite-state automata of different complexities, including parity and modular counting.
Given a random sequence of inputs, the models are trained to predict the final state, with no intermediate supervision.
The models are trained on sequences of lengths up to $40$, and we report the average validation performance on sequences of length $40$ to $256$. This setting of measuring \emph{length generalization} avoids fixed-length shortcut solutions~\cite{liu_transformers_2023}, and high accuracy indicates that the model has learned to correctly emulate the automaton.
This experimental setup exactly conforms to those used for evaluating the baseline methods~\cite{walker_2025_structured, terzic_2025_pd}, with each model having two layers\footnote{We reproduce Mamba3 with the setup of~\citet{deletang_neural_2023}, which uses 2 layers instead of 3 as in the Mamba3 paper~\cite{anonymous2025mamba3}.}.
%
%
The results shown in Table~\ref{tab:fsa-table-sum} confirm that structured sparse SSMs achieve notably higher performance compared to all alternatives. 

\begin{table*}[ht]
\centering
\caption{\textbf{Results on Synthetic State-Tracking.} Mean and standard error of validation accuracy (\%) across 5 seeds for a range of models on FSA emulation tasks. All baseline results are taken from~\cite{walker_2025_structured, terzic_2025_pd}, while the tasks were originally defined in~\cite{deletang_neural_2023}. $\dagger$ is reproduced.}
\label{tab:fsa-table-sum}
\setlength{\tabcolsep}{4pt} 
\footnotesize 
\begin{tabular}{ll cccc c}
\toprule 
\textbf{$A(u_t)$ Structure} & \textbf{Model} & \textbf{Cycle Nav.} & \textbf{Even Pairs} & \textbf{Mod. Arith.} & \textbf{Parity} & \textbf{Avg.} \\
\midrule
N/A & Transformer & 24.4 ± 0.5 & 90.4 ± 10.4 & 23.6 ± 0.7 & 52.2 ± 0.4 & 47.7 ± 2.6 \\
\midrule
$\mathbb{R}$ Diagonal & Mamba2 & 48.4 ± 2.2 & \textbf{100.0 ± 0.0} & 33.1 ± 6.6 & 54.2 ± 2.1 & 58.9 ± 1.8 \\
$\mathbb{C}$ Diagonal & Mamba3\textsuperscript{$\dagger$} & 69.0 ± 11.7 & \textbf{100.0 ± 0.0} & 46.8 ± 5.5 & 76.5 ± 7.8 & 73.0 ± 6.2 \\
DPLR & DeltaNet & 49.8 ± 4.7 & \textbf{100.0 ± 0.0} & 42.2 ± 4.8 & 57.8 ± 0.8 & 62.5 ± 1.7 \\
DPLR & DeltaNet[-1,1] & 46.7 ± 6.1 & \textbf{100.0 ± 0.0} & 66.4 ± 8.8 & \underline{97.7 ± 2.0} & 77.7 ± 2.7 \\
DPLR & Gated DeltaNet & 53.8 ± 8.8 & \textbf{100.0 ± 0.0} & 42.8 ± 8.2 & 56.5 ± 1.9 & 63.3 ± 3.0 \\
DPLR & $\text{DPLR-SLiCE}_{d_h=57,r=4}$ & 81.1 ± 16.6 & \textbf{100.0 ± 0.0} & 68.3 ± 19.3 & 91.0 ± 18.0 & 85.1 ± 13.5 \\
DPLR Product & Gated DeltaProd. & 46.3 ± 6.6 & \textbf{100.0 ± 0.0} & 78.4 ± 10.9 & 98.0 ± 1.4 & 80.7 ± 3.2 \\
Block Diagonal & $\text{BD-SLiCE}_{d_h=128,b=4}$ & 99.8 ± 0.2 & 85.9 ± 11.3 & 54.0 ± 12.5 & 95.3 ± 3.9 & 83.8 ± 6.9 \\
\midrule
Struct. Sparse & PD-SSM & \textbf{99.5 ± 0.7} & \underline{99.7 ± 0.3} & \textbf{96.2 ± 3.4} & \textbf{99.9 ± 0.1} & \textbf{98.8 ± 0.9} \\
Struct. Sparse & \textbf{\name} & \underline{92.7 ± 1.9} & 98.8 ± 0.8 & \underline{94.7 ± 2.9} & \underline{99.0 ± 1.5} & \underline{96.3 ± 1.8} \\
\midrule
N/A & Random & 20.0 & 50.0 & 20.0 & 50.0 & 35.0 \\
\bottomrule
\end{tabular}
\end{table*}

\paragraph{Mechanistic Capabilities.}
The Mechanistic Architecture Design (MAD) benchmark~\cite{poli2024mechanistic} defines 6 small-scale synthetic tasks, high performance on which has been correlated with high performance of compute-optimal trained hybrid models. 
The benchmark probes the models' capabilities to perform in-context recall, memorization, and other tasks, versions of which can be more naturally expressed as pushdown automata rather than FSAs~\cite{deletang_neural_2023}.
%
%
%
The average performance on the six tasks is reported in Table~\ref{tab:mad_results}, which shows that the Transformer is still the dominant architecture, in line with observations on related synthetic evaluations in~\cite{jelassi_repeat_2024}.  
SSMs all perform similarly, with \name's results on par with Mamba in particular.
%
%
These synthetic results indicate DeltaNet as the optimal SSM candidate for hybridization with \name\ to achieve the highest language modeling performance. We investigate this combination further in Section~\ref{subsec:hybrid}.


\textbf{Time-Series Modeling.} We additionally test \name's performance on two time-series tasks involving sequences of length over 17\,000~\cite{uea_time_series, ppg_dalia}.
%
On the UEA multivariate time-series classification dataset,~\name\ defines a new SoTA accuracy, while on the PPG-DaLiA forecasting task it also reaches comparable MSE. The results are reported in Appendix~\ref{apx:time_series}.


\begin{table}
\centering
\caption{\textbf{Results on MAD.} Accuracy (\%) on the Mechanistic Architecture Design (MAD) suite.}
\label{tab:mad_results}
\small
\setlength{\tabcolsep}{4pt} 
\begin{tabular}{lccccccc} 
\toprule
\textbf{Model}                 & \textbf{Compress} & \textbf{Fuzzy Rec.} & \textbf{IC Rec.} & \textbf{Memorize} & \textbf{Noisy Rec.} & \textbf{Sel. Copy} & \textbf{Avg.}  \\ 
\midrule
Transformer          & 51.6     & 29.8  & 94.1      & 85.2     & 86.8  & 99.6      & \textbf{74.5}     \\
\midrule
Mamba                & \textbf{52.7}     & 6.7   & 90.4      & \textbf{89.5}     & 90.1  & 86.3      & 69.3     \\
GLA      & 38.8     & 6.9   & 80.8      & 63.3     & 81.6  & 88.6      & 60.0     \\
DeltaNet             & 42.2     & \textbf{35.7}  & \textbf{100.0}     & 52.8     & \textbf{100.0} & \textbf{100.0}     & \textbf{71.8}     \\
\textbf{\name} & 47.8     & 16.7  & 97.4      & 71.9     & 96.5  & 92.2      & 70.4 \\
\bottomrule
\end{tabular}
\end{table}

\subsection{Natural Language Evaluation of Hybrid \name\ Models}\label{subsec:hybrid}
After assessing \name's performance in synthetic tasks, we evaluate its language modeling capabilities through hybridization with Mamba2, DeltaNet, as well as with self-attention.

\paragraph{State-Tracking in Natural Language.}\label{sec:nlst_results}
First, we evaluate \name\ on a natural language analogue of the state-tracking tasks, to verify that state-tracking improvements translate to the language domain. In particular, our natural language state-tracking task is based on the Box dataset~\citep{kim-schuster-2023-entity}. 
%
%
We modify the original task to include a more challenging out-of-distribution (OOD) scenario, with up to 26 state transitions, to better evaluate the capabilities of modern large architectures.
Specifically, we train on up to 6 state transitions (in-distribution, or IID), and test on up to 26 state transitions (OOD). We utilize variants of two publicly available hybrid models, namely Granite4~\citep{granite2025} at 300M, 1B, 3B, and 7B parameters, and Nemotron-H~\citep{blakeman2025nemotron} at 9B, 12B, and 30B parameters.

\begin{table}[ht]
\centering
\caption{\update{\textbf{Results on large-scale natural language state-tracking.} OOD accuracy (\%) on the Box natural language state-tracking task. Up to 4 layers of the indicated SSM are inserted into the hybrid model and trained, while the rest is kept frozen. $\dagger$ indicates that the combination could not be evaluated due to unstable training.}}
\label{tab:box_results}
\small\setlength{\tabcolsep}{4.5pt}
\begin{tabular}{lccccccc} 
\toprule
             & \multicolumn{4}{c}{Granite4}                                                                                                  & \multicolumn{3}{c}{NemotronH}                                               \\ 
\cmidrule(lr){2-5}\cmidrule(lr){6-8}
             & \multicolumn{1}{c}{300M} & \multicolumn{1}{c}{1B} & \multicolumn{1}{c}{3B} & \multicolumn{1}{c}{7B} & \multicolumn{1}{c}{9B} & \multicolumn{1}{c}{12B} & \multicolumn{1}{c}{30B}  \\
 \cmidrule(lr){2-5}\cmidrule(lr){6-8}
\update{LoRA}       & 38.08                    & 54.37                  & 56.47                  & 48.20                                     & 83.95                  & -                   & -                    \\
Mamba2       & 53.29                    & 72.81                  & 85.32                  & 87.65                                     & 79.89                  & 82.04                   & 89.14                    \\
\update{Mamba3}       & \textbf{54.92}                    & \textbf{76.44}                  & \underline{85.33}                  & 77.09                                     & $\dagger$                  & $\dagger$                   & 82.34                    \\
DeltaNet       & 49.64                    & 68.59                  & 84.12                  & \underline{88.63}                                     & \underline{82.02}                  & \underline{84.77}                   & \underline{90.45}                    \\
\textbf{\name} & \underline{53.42}           & \underline{74.95}         & \textbf{90.68}         & \textbf{89.79}           & \textbf{89.89}         & \textbf{89.25}          & \textbf{91.77}           \\
\bottomrule
\end{tabular}
\end{table}

The chosen hybrid models contain predominantly Mamba2 layers, with standard attention layers interspersed at different intervals depending on the architecture. 
\update{We leverage this design by inserting either one \name\ layer, one DeltaNet layer, one Mamba2, or one Mamba3 layer before each attention layer, for a total of up to 4 candidate layers (see Appendix~\ref{app:nlst_abl} for an ablation of the number of candidate layers).}
The setting described above allows us to keep the original model fully frozen, while training only the inserted layers to isolate their contribution to the overall performance.

\update{Table~\ref{tab:box_results} shows that \name\ provides a notable improvement to the OOD performance on this challenging task of up to $6\%$ on the Granite4 family and $10\%$ on Nemotron-H over Mamba2, and $7\%$ and $8\%$ respectively over DeltaNet. Mamba3 performs well on the smallest configurations (Granite4 300M and 1B), but \name\ surpasses it by up to 12\% on Granite4 and 9\% on Nemotron-H in larger architectures. Moreover, Mamba3 fails to converge on some models, indicating trainability issues with common layer configurations. Our results show that even very few layers of \name, less than $10\%$ of the model, are sufficient to improve the LLM's capabilities, providing SoTA accuracy at a contained additional computational cost.}

\paragraph{Language Modeling and Zero-Shot Evaluation.}\label{sec:lm_results}
\update{Finally, we test the pure language modeling performance of \name\ through hybridization with DeltaNet~\cite{yang2024parallelizing} and Gated DeltaNet, informed by the results on the MAD benchmark (see Section~\ref{sec:exp_eval}) and the natural language state-tracking task.
%
We train 100M and 230M-parameter models on 20B tokens of FineWeb-Edu~\cite{penedo2024fineweb} and evaluate them zero-shot with the lm-eval harness~\cite{eval-harness} (Table~\ref{tab:lm_results}; setup in Appx.~\ref{apx:lm_details}). 
This pretraining budget is about $10\times$ and $4\times$ the compute-optimal number of tokens~\cite{hoffmann2022chinchilla} for the 100M and 230M models, respectively.
Besides pure models, we test two hybrids with a 1:3 ratio~\cite{ren_2024_samba,granite2025,blakeman2025nemotron}: \textit{\hybridname}, DeltaNet with \name\ at every third layer, and Transformer-PD, \name\ with attention at every third layer.
At both scales, Transformer-PD attains the lowest training loss and the highest average accuracy, above the Transformer ($40.9\%$ vs.\ $39.3\%$ at 100M, $44.0\%$ vs.\ $43.7\%$ at 230M). \textit{\hybridname} has a lower training loss than DeltaNet at both scales, matches its accuracy at 100M ($39.6\%$) and surpasses it at 230M ($43.1\%$ vs.\ $42.8\%$). Pure \name\ lags mainly on LAMBADA, which requires recall from the context; both hybrids close this gap.

Informed by the results at these sizes, we also train a larger 3B model based on the Granite4 architecture, a hybrid that interleaves self-attention with SSM layers; we keep its attention layers and only exchange the SSM layers. As before, we evaluate a Gated DeltaNet model and a hybrid of \name\ and Gated DeltaNet, with \name\ at layers 1, 7, 11, 17, 21, 27, 31, and 37. Here, we also include the Mamba2 and Mamba3 models as ablations. Table~\ref{tab:lm_results_big} shows that the Gated DeltaNet-PD hybrid achieves the highest average accuracy ($59.7\%$), surpassing Gated DeltaNet ($58.1\%$), Mamba2 ($59.4\%$), and Mamba3 ($59.5\%$), while also attaining the lowest training loss and the lowest Wikitext and LAMBADA perplexities.}


\begin{table}[t]
\centering
\caption{\textbf{Results on language modeling.} Zero-shot accuracy (\%) of models trained on 20B tokens of FineWeb-Edu. FW loss: EMA of the training loss. Avg.: mean of the eight accuracies. Best and second best per scale in \textbf{bold} and \underline{underlined}. \hybridname: DeltaNet with \name\ at every third layer; Transformer-PD: \name\ with attention at every third layer.}
\label{tab:lm_results}
\label{tab:hybrid_tf}
\setlength{\tabcolsep}{2.0pt}
\footnotesize
\begin{tabular}{lccccccccccc}
\toprule
\textbf{Model} & 
\makecell{FW\\ loss $\downarrow$} & 
\makecell{LAMB.\\ ppl $\downarrow$} & 
\makecell{LAMB.\\ acc $\uparrow$} & 
\makecell{HellaS.\\ acc\_n $\uparrow$} & 
\makecell{PIQA\\ acc $\uparrow$} & 
\makecell{Arc-E\\ acc $\uparrow$} & 
\makecell{Arc-C\\ acc\_n $\uparrow$} & 
\makecell{WinoGr.\\ acc $\uparrow$} & 
\makecell{OBQA\\ acc $\uparrow$} & 
\makecell{BoolQ\\ acc $\uparrow$} & 
\makecell{Avg.\\ acc $\uparrow$} \\
\midrule
\multicolumn{12}{l}{\textit{100M Parameters}} \\
Transformer & \underline{2.68} & \textbf{81.5} & \textbf{25.9} & \underline{31.7} & 61.2 & 49.8 & 24.3 & 50.0 & \textbf{20.0} & 51.6 & 39.3 \\
DeltaNet & 2.75 & 98.8 & 23.0 & 30.9 & 59.7 & 48.6 & 24.2 & 51.4 & 17.4 & \underline{61.6} & \underline{39.6} \\
\name & 2.83 & 355.8 & 10.8 & 31.1 & \underline{61.6} & \textbf{51.6} & \textbf{25.0} & \underline{51.5} & 19.2 & 56.6 & 38.4 \\
Transformer-PD & \textbf{2.67} & \underline{82.6} & \underline{25.6} & \textbf{32.2} & \textbf{62.0} & \underline{51.1} & \underline{24.7} & \textbf{51.7} & \underline{19.6} & 60.2 & \textbf{40.9} \\
\hybridname & 2.72 & 87.7 & 24.0 & 30.8 & 61.2 & 49.8 & 24.0 & 49.3 & 16.2 & \textbf{61.8} & \underline{39.6} \\
\midrule
\multicolumn{12}{l}{\textit{230M Parameters}} \\
Transformer & \underline{2.45} & \textbf{34.8} & \underline{34.2} & \textbf{38.2} & 63.5 & 56.2 & 26.7 & \textbf{53.0} & \underline{21.2} & 56.4 & \underline{43.7} \\
DeltaNet & 2.50 & 39.8 & 30.8 & 36.4 & 63.5 & 53.7 & \underline{26.8} & 50.9 & 21.0 & \underline{59.1} & 42.8 \\
\name & 2.58 & 94.6 & 19.7 & 35.8 & 64.0 & \underline{57.1} & 26.4 & 50.4 & 21.0 & \textbf{60.3} & 41.8 \\
Transformer-PD & \textbf{2.44} & \textbf{34.8} & \textbf{34.6} & \underline{38.1} & \textbf{65.9} & \textbf{57.2} & \textbf{26.9} & \underline{52.1} & \textbf{22.8} & 54.4 & \textbf{44.0} \\
\hybridname & 2.49 & \underline{37.7} & 31.8 & 36.7 & \underline{65.3} & 56.9 & 25.9 & 51.1 & 20.0 & 57.2 & 43.1 \\
\bottomrule
\end{tabular}
\end{table}

\begin{table}
\centering
\centering
\caption{{\textbf{Results on large-scale language modeling.} Zero-shot evaluation on LM-Eval-Harness of 3B attention--SSM hybrids based on Granite4, trained on 500B tokens. The SSM layers are Gated DeltaNet, Mamba2, or Mamba3. In Gated DeltaNet-PD, 8 of the layers are \name\ and the rest Gated DeltaNet.}}
\label{tab:lm_results_big}

\setlength{\tabcolsep}{2.4pt}
\footnotesize

\resizebox{\linewidth}{!}{%
\begin{tabular}{lcccccccccccc} 
\toprule
\textbf{Model} & 
\makecell{Train\\ {loss} $\downarrow$} & 
\makecell{Wikitext\\ ppl $\downarrow$} & 
\makecell{LAMB.\\ ppl $\downarrow$} & 
\makecell{LAMB.\\ acc $\uparrow$} & 
\makecell{HellaS.\\ acc\_n $\uparrow$} & 
\makecell{PIQA\\ acc $\uparrow$} & 
\makecell{Arc-E\\ acc $\uparrow$} & 
\makecell{Arc-C\\ acc\_n $\uparrow$} & 
\makecell{WinoGr.\\ acc $\uparrow$} & 
\makecell{OBQA\\ acc $\uparrow$} & 
\makecell{BoolQ\\ acc $\uparrow$} &
\makecell{Avg.\\ acc $\uparrow$} \\
\midrule
\multicolumn{13}{l}{\textit{3B Parameters}}                                                                                                                                    \\
Gated DeltaNet           & 1.96 & 13.1                                                             & \underline{7.19}                                                     & \underline{57.6}                                                  & \underline{49.9}                                                                & 75.2                                                          & \underline{79.5}                                                           & 45.2                                                              & 62.4  & 30.6                                                           & 64.7                                                         & 58.1                        \\
\textbf{Gated DeltaNet-PD} & \textbf{1.95} & \textbf{12.9}                                                    & \textbf{6.66}                                                             & \textbf{60.0}                                                           & \textbf{50.2}                                                       & \underline{75.8}                                                          & \textbf{79.6}                                                   & \textbf{49.1}                                                      & \underline{64.2}                                                             & 31.8                                                 & 66.8                                                    & \textbf{59.7}                                                                                      \\\addlinespace[1mm]
Mamba2 & \underline{1.96} & \underline{13.1}                                                    & 7.49                                                             & 57.1                                                           & \underline{49.9}                                                       & \textbf{76.3}                                                          & \underline{79.5}                                                   & \underline{46.4}                                                      & 63.4                                                             & \textbf{35.2}                                                 & \underline{67.4}                                                    & 59.4                                                                                      \\
Mamba3 & \underline{1.96} & 13.2                                                  & 7.58                                                             & 57.0                                                           & 49.2                                                       & 75.4                                                          & 78.2                                                   & 44.3                                                      & \textbf{66.2}                                                             & \underline{34.2}                                                 & \textbf{71.3}                                                  & \underline{59.5}                                                                                      \\
\bottomrule
\end{tabular}}

\end{table}

\section{Related Work}
\label{sec:related}

LTI SSMs were proposed as scalable, high-performance alternatives to Transformers for long sequences~\cite{gu_hippo_2020,gu_combining_2021,gu_efficiently_2022}. 
While initial hybrids of gated SSMs and attention showed promise~\cite{ma2022mega, fu_hungry_2023}, the introduction of time-variant mechanisms (e.g., Mamba) notably improved practical language modeling performance~\cite{gu_mamba_2023}.
Recent studies confirm that hybrids utilizing time-variant SSMs and attention yield the strongest generalization and reasoning abilities~\citep{ren_2024_samba, lenz_2025_jamba, waleffe_2024_empirical}.
Theoretical work identifying limitations in diagonal transition matrices~\cite{merrill_illusion_2024,cirone_theoretical_2024} has spurred investigation into block-diagonal~\cite{fan_advancing_2024} and diagonal plus low-rank (DPLR) matrices~\cite{yang2024parallelizing,siems_2025_deltaproduct}. 
The latter draws inspiration from fast-weight programmers~\cite{schlag2021linear} and leverages generalized Householder matrix products for optimization~\cite{yang2024parallelizing, wy_rep_1087}.

\citet{sarrof2024expressivecapacitystatespace} show that complex diagonal SSMs are more expressive than real ones, explaining the results we obtained for Mamba3 against Mamba2 in the synthetic state-tracking task. Nonetheless, stacked complex diagonal SSMs can only emulate a limited class of automata, with depth scaling by Krohn-Rhodes (KR) complexity.
Complementing this, \citet{merrill_illusion_2024} upper-bound diagonal selective SSMs to the $\text{L-uniform }TC^0$ class, which contains all automata with solvable transformation semigroups.
DPLR matrices offer greater expressivity, capable of emulating any automaton with SSM depth scaling as a function of the automaton's complexity~\cite{siems_2025_deltaproduct}.

\section{Limitations and Conclusion}\label{sec:limitations_conclusions}

State-space models (SSMs) are increasingly utilized as the backbone of large language models, but the methods proposed so far all target an unbalanced expressivity-efficiency trade-off, leaning heavily towards efficiency.
We propose \name, a novel method that leverages structured sparse SSMs to provide training throughput comparable to that of Mamba2 while achieving FSA emulation expressivity significantly higher than that of a wide range of structured SSMs, and we show that it is an effective drop-in replacement for hybrid language models. 

At the same time, modeling complexities arise from the use of straight-through estimators, which can yield biased gradients~\cite{paulus_rao_2021}. As such, an explanation of their effectiveness on the wide range of tasks we utilized them in, in a highly discrete sequential method, is not sufficiently described by existing literature. This avenue of future research could also provide more clarity regarding the relative underperformance of pure \name\ models in the language modeling tasks.

\section*{Acknowledgment}
This work is supported by the Swiss National Science Foundation (SNSF), grant 10002666.

\bibliography{bibliography}
\bibliographystyle{icml2026}

\newpage
\appendix
\makeatletter\setlength{\@fptop}{0pt}\makeatother
\onecolumn

\section{Additional Experiments}
\label{sec:apx_experiments}

\subsection{Time-Series}\label{apx:time_series}

\textbf{Classification} We next evaluate our model on a subset of the University of East Anglia Multivariate Time-Series Classification Archive (UEA-MTSCA)~\citep{uea_time_series}, extending on the results from~\citet{walker_2024_logncde, rusch2025oscillatory}. We consider six tasks selected due to their long sequence lengths, which range from around 400 to over 17\,000. Conforming to the evaluation methodology and data splits defined by~\citet{rusch2025oscillatory, walker_2024_logncde}, we report the average and standard deviation of the test accuracy across five random initializations, with the hyperparameter grid as defined in~\citet{rusch2025oscillatory}. The results reported in Table~\ref{tab:time_series} show that \name\ is SoTA on average with comparable standard deviation.
\begin{table}[h!]
\caption{\textbf{Results on time-series classification.} Mean and standard deviation of test accuracies (\%) across 5 seeds on selected long-sequence UEA time-series classification datasets.}
\label{tab:time_series}
\centering
\footnotesize
\resizebox{\linewidth}{!}{%
\begin{tabular}{lccccccc}
\toprule
{\textbf{Model}} & 
\textbf{Worms} & 
\textbf{SCP1} & 
\textbf{SCP2} & 
\textbf{Ethanol} & 
\textbf{Heartbeat} &
\textbf{Motor} &
{\textbf{Average}} \\
\midrule
NCDE & 75.0 ± 3.9 & 79.8 ± 5.6 & 53.0 ± 2.8 & 29.9 ± 6.5 & 73.9 ± 2.6 & 49.5 ± 2.8 & 60.2 ± 1.76 \\
Log-NCDE & 85.6 ± 5.1 & 83.1 ± 2.8 & 53.7 ± 4.1 & 34.4 ± 6.4 & 75.2 ± 4.6 & 53.7 ± 5.3 & 64.3 ± 1.99 \\
LRU & {87.8 ± 2.8} & 82.6 ± 3.4 & 51.2 ± 3.6 & 21.5 ± 2.1 & \underline{78.4 ± 6.7} & 48.4 ± 5.0 & 61.7 ± 1.72 \\
S6 & 85.0 ± 16.1 & 82.8 ± 2.7 & 49.9 ± 9.4 & 26.4 ± 6.4 & 76.5 ± 8.3 & 51.3 ± 4.7 & 62.0 ± 3.68 \\
Mamba & 70.9 ± 15.8 & 80.7 ± 1.4 & 48.2 ± 3.9 & 27.9 ± 4.5 & 76.2 ± 3.8 & 47.7 ± 4.5 & 58.6 ± 2.99 \\
LinOSS-IMEX  & 80.0 ± 2.7 & {87.5 ± 4.0} & \textbf{58.9 ± 8.1} & 29.9 ± 1.0 & 75.5 ± 4.3 & \underline{57.9 ± 5.3} & {65.0 ± 1.95} \\
LinOSS-IM & \textbf{95.0 ± 4.4} & \underline{87.8 ± 2.6} & \underline{58.2 ± 6.9} & {29.9 ± 0.6} & 75.8 ± 3.7 & \textbf{60.0 ± 7.5} & \underline{67.8 ± 2.00} \\
PD-SSM & 90.0 ± 5.7 & 80.9 ± 2.0 & {56.1 ± 8.6} & \underline{34.7 ± 4.0} & \textbf{80.0 ± 2.6} & \textbf{60.0 ± 3.7} & 67.0 ± 2.02 \\
\textbf{\name} & \underline{93.8 ± 1.4} & 86.1 ± 2.8 & 56.1 ± 3.1 & \textbf{43.7 ± 2.9} & 77.2 ± 6.6 & 57.9 ± 6.0 & \textbf{69.1 ± 2.02} \\
\bottomrule
\end{tabular}}
\end{table}

\textbf{Forecasting} We further evaluate the performance of \name\ in the context of extensive long-range dependencies on the PPG-DaLiA dataset~\cite{ppg_dalia}, a multivariate time series regression benchmark developed for heart rate estimation. The dataset comprises approximately 150 minutes of data per subject across 15 individuals, recorded at a maximum sampling frequency of 128 Hz. The input features consist of six distinct channels: blood volume pulse, electrodermal activity, body temperature, and triaxial acceleration.
Following the methodology from~\cite{walker_2024_logncde}, the data for each subject were partitioned into training, validation, and test sets using a 70/15/15 ratio. Post-partitioning, a sliding window of 49\,920 samples was implemented with a stride of 4\,992.
Empirical results shown in Table~\ref{tab:ppg} demonstrate that our proposed method reaches a near SoTA score in terms of test mean-squared error (MSE), rivaled only by the LinOSS methods~\cite{rusch2025oscillatory}. 

\begin{table}[ht]
    \caption{\textbf{Results on time-series forecasting}. Mean squared error and standard deviation on the PPG-DaLiA dataset.}
    \label{tab:ppg}
    \centering
    \footnotesize
    \setlength{\tabcolsep}{3.5pt}
    \resizebox{\linewidth}{!}{%
    \begin{tabular}{lcccccccc}
        \toprule
        Model & NCDE & Log-NCDE & LRU & S6 & Mamba & LinOSS-IMEX & LinOSS-IM & \textbf{\name} \\
        \midrule
        MSE $\times 10^{-2}$ & $13.54 \pm 0.69$ & $9.56 \pm 0.59$ & $12.17 \pm 0.49$ & $12.88 \pm 2.05$ & $10.65 \pm 2.20$ & $7.5 \pm 0.46$ & $\mathbf{6.4 \pm 0.23}$ & \underline{$6.5 \pm 0.35$} \\
        \bottomrule
    \end{tabular}}
\end{table}

\subsection{Natural language state tracking}\label{app:nlst_abl}

Our natural language state-tracking task is based on the Box dataset~\citep{kim-schuster-2023-entity}. 
Here, a series of objects and boxes is described in plain English. At each step, an object might be either inserted or removed from a box, or transferred between boxes. 
The task is to predict, again in plain English, the content of each box at each step.
The most natural setting for the evaluation of the natural language state tracking task on the hybrid models we have chosen (Granite4 and Nemotron-H) suggests the insertion of 4 additional SSM layers to be finetuned. The full results are presented in Table \ref{tab:box_results_full}.

\begin{table}[ht]
\centering
\caption{\textbf{Results on large-scale natural language state-tracking.} Accuracy (\%) on the Box natural language state-tracking task. Up to 4 layers of the indicated SSM are inserted into the model and trained, while the rest is kept frozen.}
\label{tab:box_results_full}
\footnotesize
\setlength{\tabcolsep}{2.8pt}
\begin{minipage}[t]{0.49\linewidth}
\centering
\begin{tabular}{l l c c}
\toprule
\textbf{Model} & \textbf{SSM} & {IID} & {OOD} \\
\cmidrule(lr){1-1}\cmidrule(lr){2-2} \cmidrule(lr){3-4}
\multirow{3}{*}{Granite4-300M} & Mamba2 & 83.77    & \underline{53.29} \\
                                & DeltaNet  & 82.42    & 49.64 \\
                                & \textbf{\name}  & 82.85    & \textbf{53.42} \\\cmidrule(lr){1-1}\cmidrule(lr){2-2} \cmidrule(lr){3-4}
\multirow{3}{*}{Granite4-1B}   & Mamba2 & 96.51    & \underline{72.81} \\
                                & DeltaNet  & 95.72    & 68.59 \\
                                & \textbf{\name}  & 95.92    & \textbf{74.95} \\
                                \cmidrule(lr){1-1}\cmidrule(lr){2-2} \cmidrule(lr){3-4}
\multirow{3}{*}{Granite4-3B}    & Mamba2 & 98.41    & \underline{85.32} \\
& DeltaNet  & 97.53    & 84.12 \\
                                & \textbf{\name}  & 98.53    & \textbf{90.68} \\
                                \cmidrule(lr){1-1}\cmidrule(lr){2-2} \cmidrule(lr){3-4}
\multirow{3}{*}{Granite4-7B}    & Mamba2 & 99.23    & 87.65 \\
                                & DeltaNet  & 99.12    & \underline{88.63} \\
                                & \textbf{\name}  & 98.39    & \textbf{89.79} \\
\bottomrule
\end{tabular}
\end{minipage}\hfill
\begin{minipage}[t]{0.49\linewidth}
\centering
\begin{tabular}{l l c c}
\toprule
\textbf{Model} & \textbf{SSM} & {IID} & {OOD} \\
\cmidrule(lr){1-1}\cmidrule(lr){2-2} \cmidrule(lr){3-4}
\multirow{3}{*}{NemotronH-9B}  & Mamba2 & 97.30 & 79.89 \\
                                & DeltaNet  & 96.18 & \underline{82.02} \\
                                & \textbf{\name}  & 99.71 & \textbf{89.89} \\
                                \cmidrule(lr){1-1}\cmidrule(lr){2-2} \cmidrule(lr){3-4}
\multirow{3}{*}{NemotronH-12B}  & Mamba2 & 96.59 & 82.04 \\
                                & DeltaNet  & 96.22 & \underline{84.77} \\
                                & \textbf{\name}  & 99.71 & \textbf{89.25} \\
                                \cmidrule(lr){1-1}\cmidrule(lr){2-2} \cmidrule(lr){3-4}
\multirow{3}{*}{NemotronH-30B}  & Mamba2 & 99.24 & 89.14 \\
& DeltaNet  & 98.02 & \underline{90.45} \\
                                & \textbf{\name}  & 99.41 & \textbf{91.77} \\
\bottomrule
\end{tabular}
\end{minipage}
\end{table}

\textbf{Number of inserted layers.} We have seen in Section~\ref{sec:nlst_results} that this setup is both effective at solving the task and computationally efficient. Nonetheless, we also vary the number and position of the inserted layers using Granite4-3B to investigate the effectiveness of \name\ and Mamba2. For each number of layers, we report in Table~\ref{tab:box_layers_results} the average performance across all possible candidate positions. Our results show that, except for a single layer, \name\ always performs better than Mamba2. Moreover, Mamba2 already plateaus at 2 additional layers, while \name\ is able to take advantage of all 4 layers.

\begin{table}[h]
\footnotesize
\setlength{\tabcolsep}{4pt}
\begin{minipage}[t]{0.47\linewidth}
\centering
\caption{\textbf{Number of inserted layers on natural language state tracking.} Accuracy (\%) on the Box natural language state-tracking task. 1 to 4 layers of the indicated SSM are inserted into Granite4-3B and trained, while the rest is kept frozen.}
\label{tab:box_layers_results}
\begin{tabular}{l l c c}
\toprule
\textbf{Add. layers} & \textbf{SSM} & {IID} & {OOD} \\
\midrule
\multirow{2}{*}{4}   & Mamba2 & 98.41    & 85.32 \\
    & \name  & 98.53    & \textbf{90.68} \\
\midrule
\multirow{2}{*}{3}   & Mamba2 & 98.30    & 85.68 \\
    & \name  & 97.93    & \textbf{86.05} \\
\midrule
\multirow{2}{*}{2}   & Mamba2 & 98.22    & 85.55 \\
    & \name  & 98.09    & \textbf{86.74} \\
\midrule
\multirow{2}{*}{1}   & Mamba2 & 96.44 & \textbf{82.83} \\
    & \name  & 96.55 & 80.84 \\
\bottomrule
\end{tabular}
\end{minipage}\hfill
\begin{minipage}[t]{0.49\linewidth}
\centering
\caption{\textbf{Effectiveness of \name{}.} \name{}\ or Mamba2 are inserted into Granite4-3B and evaluated on the Box natural language state-tracking task.}
\label{tab:box_int_results}
\begin{tabular}{llcc}
\toprule
\makecell{\textbf{No.}\\ \textbf{\name{}}} & \makecell{\textbf{No.}\\ \textbf{Mamba2}} & {IID} & {OOD} \\
\midrule
1 layer & 3 layers & 98.47 & 87.79 \\
2 layers & 2 layers & 98.15 & 88.63 \\
3 layers & 1 layer & \textbf{98.64} & 88.94 \\
4 layers & 0 layers & 98.53 & \textbf{90.68} \\
\bottomrule
\end{tabular}

\vspace{0.9em}
\caption{\textbf{Dictionary size on natural language state tracking.} Accuracy (\%) on the Box natural language state-tracking task. The dictionary size (K) of the 4 inserted PD-SSM layers into Nemotron-H 9B is varied from 8 to 256.}
\label{tab:box_k_results}
\begin{tabular}{l c c}
\toprule
\textbf{Dictionary size (K)} & {IID} & {OOD} \\
\midrule
8 & \textbf{99.71}    & 89.89 \\
16  & 99.49    & 89.52 \\
32  & 99.52 & 92.44 \\
64  & 99.60 & 92.90 \\
128  & 99.68 & 91.10 \\
256  & 99.70 & \textbf{94.44} \\
\bottomrule
\end{tabular}
\end{minipage}
\end{table}

Finally, we evaluate the effectiveness of \name\ and Mamba2 when both types of layers are inserted and finetuned in the Granite4-3B model. Specifically, at each of the 4 candidate positions, we insert either a \name{}\ or a Mamba2 layer, leading to 3 possible intermediate configurations: (1) 1 \name{}\ layer followed by 3 Mamba2 layers, (2) 2 layers of each type, and (3) 1 Mamba2 layer followed by 3 \name{}\ layers. The results are shown in Table~\ref{tab:box_int_results}. Even 1 \name{}\ layer already improves performance by 3\%, with 2 and 3 \name{}\ layers providing a 4\% uplift even when used in conjunction with Mamba2 layers. Still, 4 \name{}\ layers provide the best performance.

\textbf{Dictionary size.} As with the synthetic state-tracking tasks, we also vary here the dictionary size to verify that performance improves as the model's expressiveness increases. Indeed, Table~\ref{tab:box_k_results} confirms that \name's performance increases with K.

\section{Experimental Setup}\label{apx:exp_setup}

\subsection{Synthetic State Tracking}

As previously mentioned, on the 4 synthetic state tracking tasks from~\cite{deletang_neural_2023}\footnote{Generative code released at \url{https://github.com/google-deepmind/neural_networks_chomsky_hierarchy} under Apache 2.0 license.}, we constrain ourselves to the experimental hyperparameter choices made in~\cite{walker_2025_structured}. Concretely, all models consist of two layers; the learning rate scheduler comprises a linear warmup followed by cosine decay with a maximal learning rate of $0.002$. We use Adam with default parameters. The models are trained for $100{,}000$ steps with a batch size of $256$. In our main experimental results, we used a model with total state size equal to the embedding size of $128$ and 4 heads. In this set of experiments, we used the complex-valued version of \name{} with a linear readout.

\subsection{Natural Language State Tracking}

We modify the Box dataset~\citep{kim-schuster-2023-entity}\footnote{Generative code provided in \url{https://github.com/sebschu/entity-tracking-lms}.} to achieve a uniform distribution of samples across all lengths, both IID and OOD. This ensures that all lengths are appropriately represented both during training and testing, and allows for proper length generalization evaluation of large models. Moreover, we also increase the length of the testing set up to a sequence length of 26, and train up to a sequence length of 6.

Given the large-scale nature of the models tested, on the natural language state-tracking we vary the learning schedule based on the specific model to obtain the best evaluation performance. We use Adam with 0.1 weight decay in all instances. All models are trained and tested on 8$\times$Nvidia A100 GPUs, requiring from 5 to 50 hours to complete the pipeline.

On Granite4-300M, we use a constant learning schedule with a learning rate of 0.001 and 10\% of the steps used as a warmup. The same goes for Granite4-1B. With Granite4-3B and Granite4-7B, we lower the learning rate to 0.0005 and use a cosine learning schedule with 10\% of the steps used as warmup and 0.1 decay factor. Finally, for the NemotronH family we use a learning rate of 0.0001 and the same cosine learning schedule as Granite4-7B.

\subsection{MAD}

The mechanistic architecture design benchmark~\cite{madlab_github}\footnote{Generative code released under MIT license.} defined standard hyperparameter grid search containing three learning rate values $\{1e-4, 5e-4, 1e-3\}$ and two weight decay values $\{0.0, 0.1\}$. In this set of experiments, we used the real-valued version of \name{} with a nonlinear readout.

\subsection{Time-Series Results}
We run experiments on the UEA Multivariate Time Series Classification~\cite{uea_time_series}\footnote{Released at \url{https://www.timeseriesclassification.com/} under GPL-3.0 license.} conforming to the hyperparameter grid defined by~\cite{walker_2024_logncde}, which has subsequently been used to expand upon this set of results in at least the following papers:~\cite{walker_2025_structured, rusch2025oscillatory, terzic_2025_pd}. We used Adam with the default parameters. The grid search consists of the following model hyperparameters: learning rate in $\{1e{-5}, 1e{-4}, 1e{-3}\}$, number of layers in $\{2, 4, 6\}$, state dimension in $\{16, 64, 256\}$ and embedding dimension in $\{16, 64, 128\}$. 

In this set of experiments, we used the real-valued version of \name{} with a nonlinear readout.

\subsection{Efficiency Benchmark}\label{apx:efficiency}

Each model stacks $n$ pre-norm blocks (mixer, SwiGLU MLP) of width $d$, with $(d, n) \in \{(768, 12), (1536, 24), (2048, 28), (3072, 44)\}$, a 32k vocabulary and untied embeddings. Models are parameter-matched to within $1\%$ through the MLP width (\name, DeltaNet) or the mixer width (Mamba2, Gated DeltaNet). Training: one full step (forward, backward, AdamW) at micro-batch $4 \times 2048$ in bfloat16 autocast, median of 15 steps. Decoding: bfloat16, eager PyTorch, median of 128 greedy steps after a 2048-token prompt. Executed on an Nvidia H200. For complete results, see Table~\ref{tab:lm_efficiency}.

\begin{table}[h]
\centering
\small
\caption{\textbf{Efficiency of parameter-matched language models.} Training throughput (k tokens/s) and peak memory (GiB); decoding throughput at batch size 16 (tokens/s).}
\label{tab:lm_efficiency}
\setlength{\tabcolsep}{3.6pt}
\begin{tabular}{l rrr rrr rrr rrr}
\toprule
& \multicolumn{3}{c}{0.1B} & \multicolumn{3}{c}{0.5B} & \multicolumn{3}{c}{1B} & \multicolumn{3}{c}{3B} \\
\cmidrule(lr){2-4} \cmidrule(lr){5-7} \cmidrule(lr){8-10} \cmidrule(lr){11-13}
& train & mem & dec & train & mem & dec & train & mem & dec & train & mem & dec \\
\midrule
Mamba2 & 153.0 & \textbf{3.50} & \textbf{1874} & \textbf{65.2} & \textbf{14.16} & \textbf{939} & 38.7 & \textbf{24.37} & \textbf{802} & 10.2 & \textbf{67.89} & \textbf{519} \\
DeltaNet & 137.2 & 4.54 & 1268 & 62.3 & 18.50 & 621 & \textbf{44.1} & 30.99 & 543 & \textbf{17.0} & 83.63 & 331 \\
Gated DeltaNet & 137.3 & 4.88 & 1377 & 60.9 & 18.92 & 683 & 43.2 & 31.64 & 588 & 16.6 & 85.16 & 367 \\
\name & \textbf{178.5} & 4.28 & 1362 & 64.9 & 17.51 & 671 & 41.6 & 29.47 & 581 & 16.0 & 80.12 & 358 \\
\bottomrule
\end{tabular}
\end{table}

\subsection{Language Model Experiment Details}\label{apx:lm_details}

All models were trained on 20B tokens of the FineWeb-Edu sample-100BT~\cite{penedo2024fineweb}\footnote{Released at \url{https://huggingface.co/datasets/HuggingFaceFW/fineweb-edu} under ODC-By 1.0 license.} with the TinyLlama tokenizer and a context length of 2048, using 12 layers (100M, width 768) or 24 layers (230M, width 1024), with MLP sizes chosen to match the parameter count. Training uses AdamW (learning rate $3 \times 10^{-4}$, weight decay 0.1, gradient clipping 1.0), 524k tokens per step, a linear warmup over 1B tokens and cosine decay to $10^{-5}$, in bfloat16 on 4/8$\times$Nvidia H200 GPUs. The \name\ dictionaries are initialized as the identity plus Gaussian noise, such that each is diagonal with probability at least 0.9. In this set of experiments, we used the real-valued version of \name{}. The reported training loss is an exponential moving average of the per-step loss (span 500 steps) at the end of training.

We utilize the LM Evaluation Harness~\cite{eval-harness}\footnote{Code at \url{https://github.com/EleutherAI/lm-evaluation-harness} under MIT license.} to test the zero-shot language modeling capabilities of our pretrained models. Specifically, we report perplexity and accuracy on LAMBADA (OpenAI version)~\cite{paperno-etal-2016-lambada}; normalized accuracy on HellaSwag~\cite{zellers-etal-2019-hellaswag} and Arc-Challenge~\cite{clark2018think}; and standard accuracy on PIQA~\cite{bisk2020piqa}, Arc-Easy~\cite{clark2018think}, WinoGrande~\cite{sakaguchi2019winogrande}, OpenBookQA~\cite{OpenBookQA2018}, and BoolQ~\cite{clark2019boolq}. 

Training took between 6 and 12 hours for the 100M models and between 10 and 21 hours for the 230M models on 4 resp. 8 H200 GPUs.

\section{Deriving the Surrogate Gradient Expressions}\label{apx:proof}

Recall the transition equation $x_t=P_tD_tx_{t-1}+Bu_t$, where we set $P_t=P(u_t)$ and $D_t=D(u_t)$ for brevity.

\paragraph{The Gradient with Respect to the Dictionary Matrices}

To compute the gradient with respect to the dictionary matrices $M_k$, we will separately compute $\frac{\partial \ell}{\partial P_t}$ and $\frac{\partial P_t}{\partial M_k}$. 

Since we assumed access to the intermediate derivatives $\frac{\partial \ell}{\partial x_t}$, we can immediately obtain the first factor as
\begin{equation}
    \frac{\partial \ell}{\partial P_t} = \frac{\partial \ell}{\partial x_t} (D_t x_{t-1})^T
\end{equation}

To find the total gradient for a specific dictionary matrix $M_k$, we sum the gradients across all time steps $t$ where $M_k$ was actively selected ($k^*(u_t)=k$) and apply the chain rule:
\begin{align}
    \frac{\partial \ell}{\partial M_k} &= \sum_{t:k^*(u_t)=k} \frac{\partial \ell}{\partial P_t} \frac{\partial P_t}{\partial M_k^{\text{sparse}}} \frac{\partial M_k^{\text{sparse}}}{\partial M_k} \\
    &\approx \sum_{t:k^*(u_t)=k}\Big(\frac{\partial \ell}{\partial x_t}\;(D_t x_{t-1})^T \Big) \frac{\partial \text{softmax}_\tau(M_k)}{\partial M_k} \label{eq:grad_M}
\end{align}

\paragraph{The Gradient with Respect to the Selection Weights}

Let $v_t$ be the one-hot representation of the discrete routing choice $k^*(u_t) = \text{argmax}(k(u_t))$. We relax this to a soft probability distribution: $v_t^{\text{soft}} = \text{softmax}_\tau(k(u_t))$. 

Instead of computing the exact gradient for all unselected elements, the soft-routing surrogate approximation uses the activated forward state. Applying the chain rule alongside the gradient of the softmax yields:
\begin{equation}
    \frac{\partial \ell}{\partial k(u_t)} = \frac{\partial \ell}{\partial P_t} \frac{\partial P_t}{\partial v_t} \frac{\partial v_t^{\text{soft}}}{\partial k(u_t)} \approx \Big(\frac{\partial \ell}{\partial x_t} (P_tD_tx_{t-1})^T\Big)\frac{\partial \text{softmax}_\tau(k(u_t))}{\partial k(u_t)} \label{eq:grad_k}
\end{equation}

We explicitly compute this and the preceding expression using dedicated Triton/CUDA kernels, and chain them with the remaining network by relying on the PyTorch autodifferentiation framework.

As the temperature $\tau \rightarrow 0$, the continuous $\text{softmax}_\tau(\cdot)$ converges to the exact $\text{argmax}(\cdot)$ operation. Consequently, the soft relaxation perfectly mimics the hard selections made in the forward pass, causing the surrogate gradient expressions (\ref{eq:grad_M}) and (\ref{eq:grad_k}) to become exact.

\section{\name\ Expressivity Guarantee}
\label{sec:apx_proof_expressivity}

The expressivity guarantee is a simple restatement of the results obtained in~\cite{terzic_2025_pd}, with the construction previously described in earlier sources~\cite{straubing_book, liu_transformers_2023}.
Denote a deterministic finite-state automaton by a 5-tuple $(Q, \Sigma, \delta, q_{\text{init}}, F)$ where $Q$ is a finite set of states, $\Sigma$ is a finite input alphabet, $\delta : Q \times \Sigma \rightarrow Q$ is the state transition function, $q_{\text{init}} \in Q$ is a fixed initial state, and $F \subseteq Q$ is the set of accepting states. 

Any (deterministic) FSA can be mapped to a time-variant SSM as follows.
Encode each $q \in Q$ using $enc: Q \rightarrow\mathbb{R}^{|Q|}$ such that the encodings of different states are different one-hot vectors.
Given such an encoding of states, we can map the state transition function $\delta : Q \times \Sigma \rightarrow Q$ to the state transition matrices $A(u_t)$
via $A(\sigma)={\scriptstyle\sum\nolimits}_{q \in Q}\ enc(\delta(q, \sigma))\cdot enc(q)^T$.
Due to the one-hot structure of the state encodings, this exactly corresponds to a binary \name\ matrix, and depending on the input, a single such matrix is selected.
This requires that the dictionary size of \name\ in general be set to $K=|\Sigma|$.
If we furthermore set, in the general SSM equations shown in~(\ref{eq_slssm}), $B(u_t) =0$, $D(u_t) = 0$, and $\psi = id$, the SSM state always exactly matches the one-hot encoding of the corresponding FSA state.
Clearly, then, the set of accepting final states can be encoded in a linear readout vector of size $N$, which is defined as $C(u_t)=C=\sum_{q\in F}enc(q)^T \in \mathbb{R}^{1 \times N}$.

\section{Implementation}
\label{sec:apx_implementation}

\begin{figure}[H]
\centering
\includegraphics[width=\linewidth]{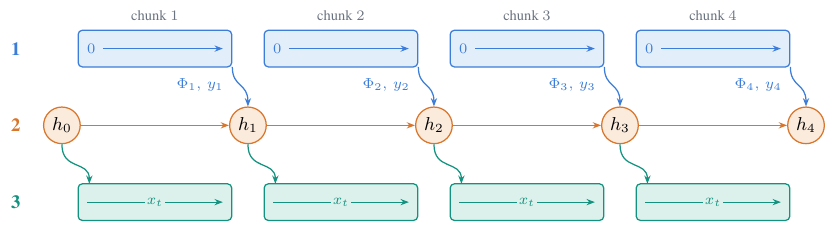}
\caption{\textbf{Three-phase chunked recurrence.} (1) Each chunk is scanned from a zero state, giving its summary $(\Phi_c, y_c)$. (2) A short sequential scan turns the summaries into the chunk-boundary states $h_c$. (3) Each chunk replays its recurrence from $h_{c-1}$ and writes $x_t$. Phases 1 and 3 run in parallel over chunks.}
\label{fig:three_phase}
\end{figure}

\name\ computes the recurrence $x_t = A_t x_{t-1} + b_t$, with $A_t = P_t D_t$, in three phases (Figure~\ref{fig:three_phase}, Algorithm~\ref{alg:three_phase}). The sequence of length $L$ is split into $C$ chunks of $\tau$ steps.
(1)~Each chunk is summarized, in parallel, by the product $\Phi_c$ of its transition matrices and the state $y_c$ it reaches from a zero initial state. Since a product of PD matrices is again a PD matrix, $\Phi_c$ takes only $\mathcal{O}(N)$ memory.
(2)~A sequential scan over the $C$ summaries gives the state $h_c$ at the end of every chunk.
(3)~Each chunk replays its recurrence from $h_{c-1}$, in parallel, and writes the states $x_t$.
Every step is a gather and a scaling of an $N$-dimensional vector. 

\begin{algorithm}[H]
\small
\caption{Three-phase chunked recurrence}
\label{alg:three_phase}
\begin{algorithmic}[1]
\Require transitions $A_t = P_t D_t$ and inputs $b_t$ for $t = 1, \dots, L$; chunk length $\tau$
\Ensure states $x_t = A_t x_{t-1} + b_t$ for $t = 1, \dots, L$, with $x_0 = 0$
\State Split $1, \dots, L$ into chunks $\mathcal{T}_1, \dots, \mathcal{T}_C$ of $\tau$ consecutive steps
\Statex \textcolor{gray}{\textit{Phase 1: chunk summaries}}
\For{$c = 1, \dots, C$ \textbf{in parallel}}
    \State $\Phi \gets I$, \ $y \gets 0$
    \For{$t \in \mathcal{T}_c$}
        \State $\Phi \gets A_t \Phi$, \ $y \gets A_t y + b_t$
    \EndFor
    \State $(\Phi_c, y_c) \gets (\Phi, y)$ \Comment{$\Phi_c$ is a PD matrix}
\EndFor
\Statex \textcolor{gray}{\textit{Phase 2: chunk-boundary states}}
\State $h_0 \gets 0$
\For{$c = 1, \dots, C$}
    \State $h_c \gets \Phi_c h_{c-1} + y_c$ \Comment{state after chunk $c$}
\EndFor
\Statex \textcolor{gray}{\textit{Phase 3: outputs}}
\For{$c = 1, \dots, C$ \textbf{in parallel}}
    \State $x \gets h_{c-1}$
    \For{$t \in \mathcal{T}_c$}
        \State $x \gets A_t x + b_t$, \ $x_t \gets x$
    \EndFor
\EndFor
\end{algorithmic}
\end{algorithm}

\end{document}